\documentclass{article} 
\usepackage{iclr2027_conference,times}

\usepackage{amsmath,amsfonts,bm}

\def\eqref#1{equation~\ref{#1}}

\def\1{\bm{1}}

\def\vx{{\bm{x}}}

\DeclareMathAlphabet{\mathsfit}{\encodingdefault}{\sfdefault}{m}{sl}
\SetMathAlphabet{\mathsfit}{bold}{\encodingdefault}{\sfdefault}{bx}{n}

\usepackage{hyperref}
\usepackage{multirow}
\usepackage{graphicx}
\usepackage{tikz}
\usepackage{natbib}
\usepackage[utf8]{inputenc}
\usepackage[T1]{fontenc}
\usepackage{microtype}

\usepackage{amssymb}
\usepackage{mathtools}

\usepackage{graphicx}
\graphicspath{{figures/}}          
\usepackage[labelfont=bf]{caption}
\usepackage{subcaption}
\usepackage{placeins}     

\usepackage{tikz}
\usetikzlibrary{calc,positioning,patterns,arrows.meta,decorations.pathreplacing}

\usepackage{booktabs}
\usepackage{multirow}
\usepackage{array}

\usepackage{algorithm}
\usepackage{algpseudocode}
\usepackage{float}

\usepackage[dvipsnames]{xcolor}
\hypersetup{colorlinks=true, linkcolor=BrickRed, citecolor=MidnightBlue, urlcolor=MidnightBlue}
\usepackage{url}
\usepackage[capitalise,noabbrev]{cleveref}

\usepackage{xspace}
\usepackage[colorinlistoftodos,textsize=scriptsize]{todonotes}

\usetikzlibrary{
    positioning,
    arrows.meta,
    calc,
    fit,
    backgrounds,
    matrix,
    decorations.pathreplacing
}

\definecolor{targetc}{RGB}{70,130,180}
\definecolor{refc}{RGB}{220,120,60}
\definecolor{textc}{RGB}{100,160,100}
\definecolor{blockc}{RGB}{235,235,245}
\definecolor{highlightc}{RGB}{255,220,120}
\definecolor{softblue}{RGB}{220,235,255}
\definecolor{softorange}{RGB}{255,235,220}
\definecolor{softgreen}{RGB}{230,245,230}
\makeatletter
\newcommand{\citecolortext}[1]{\textcolor{\@citecolor}{#1}}
\makeatother
\tikzset{
    >={Latex[length=2mm]},
    box/.style={
        draw,
        rounded corners,
        thick,
        align=center,
        minimum height=8mm,
        minimum width=16mm,
        fill=white
    },
    smallbox/.style={
        draw,
        rounded corners,
        thick,
        align=center,
        minimum height=6mm,
        minimum width=10mm,
        fill=white
    },
    stage/.style={
        draw,
        rounded corners,
        thick,
        fill=blockc,
        minimum height=10mm,
        minimum width=28mm,
        align=center
    },
    img/.style={
        draw,
        thick,
        minimum width=15mm,
        minimum height=15mm,
        align=center
    },
    tok/.style={
        draw,
        thick,
        minimum width=5mm,
        minimum height=7mm,
        fill=white
    },
    attnblock/.style={
        draw,
        thick,
        rounded corners,
        fill=blockc,
        minimum width=34mm,
        minimum height=18mm,
        align=center
    },
    lab/.style={font=\small},
    sml/.style={font=\scriptsize},
}

\newcommand{\kk}[1]{}

\newcommand{\method}{CaD\xspace}
\newcommand{\decontext}{DeContext\xspace}
\newcommand{\kontext}{Flux Kontext\xspace}
\newcommand{\fibo}{FIBO\xspace}

\newcommand{\diffpgd}{Diff-PGD\xspace}

\title{Copying as Defense: Protecting Images from Editing by Eliciting Model Copying}

\author{
Salma Abdel Magid$^{1}$, Sanghyuk Chun$^{1}$, Eric Slyman$^{2}$, Kushal Kafle$^{2}$, Olga Russakovsky$^{1}$ \\
$^{1}$Princeton University, $^{2}$Adobe Research
}

\newcommand{\secondbest}[1]{\textcolor[RGB]{90,150,220}{#1}}
\iclrfinalcopy 
\begin{document}

\maketitle

\begin{abstract}
Image editors built on diffusion transformers (DiTs), such as \kontext, make it possible to turn a single photograph into a convincing manipulated image, including harmful and non-consensual edits. Proactive defenses aim to prevent such manipulations by adding an imperceptible perturbation before an image is shared. Existing defenses for DiT-based in-context editors take an intuitive approach: make the output attend \emph{less} to the protected reference image. But this leaves the model free to follow the edit instruction using the text and its own generated tokens, often producing a new, weakly grounded image rather than preventing the edit altogether. We find that the opposite strategy is more effective. Instead of leading the model to ignore the reference, we encourage it to \emph{copy it}. Through controlled attention interventions, we show that concentrating target-token attention on the reference image in only a few early blocks is sufficient to collapse the editing behavior, causing the model to reproduce its input, often as a lossy copy, rather than execute the requested transformation. 
Based on this insight, we introduce Copying-as-Defense (\method), which maximizes per-token target-to-reference attention mass in early transformer blocks. Since the attention rows are normalized, increasing reference attention suppresses the attention allocated to the text and generated canvas. Restricting the objective to early blocks also lets us truncate backpropagation path, keeping optimization cheap. 
Across editors and benchmarks, CaD substantially suppresses instruction-following. It reduces average edit success from $81.4\%$ on unprotected images to $24.0\%$, a $70.6\%$ relative reduction. Compared with DeContext, the strongest baseline on average at $43.0\%$ edit success, CaD further reduces edit success by $44.3\%$. CaD is also substantially cheaper to optimize, running roughly $1.8$--$2.0\times$ faster per step than comparable optimization-based methods.

\end{abstract}

\section{Introduction}
A photograph posted online can now be transformed into a convincing manipulated image with a single natural-language instruction. The same capability that enables benign editing also lowers the barrier to harmful uses, including non-consensual sexualized imagery, impersonation, misinformation, fraud, and copyright misuse 
\citep{bodybench,nbc2026grok,marchal2024misuse}. One response has been to protect an image before it is shared by adding a small, imperceptible perturbation designed to make downstream edits fail. This idea predates modern diffusion editors, with work disrupting GAN-based facial manipulation systems~\citep{ruiz2020disrupting}, and has since expanded to diffusion-based editing~\citep{salman2023photoguard,choi2025diffusionguard}, unauthorized personalization \cite{marchal2024misuse}, and artistic imitation~\citep{liang2023advdm,shan2023glaze,honig2025adversarial}. As generative architectures evolve, however, the mechanism a defense must disrupt evolves with them.

That shift is especially important for the latest generation of diffusion transformers (DiTs)~\citep{peebles2023dit,chen2024pixartalpha,ma2024sit}. A DiT replaces the convolutional U-Net backbone with a transformer operating on latent patches~\citep{peebles2023dit}. Multimodal DiTs (MMDiT), introduced with Stable Diffusion~3, instead use separate weights for text and image tokens and let information flow in both directions between them~\citep{esser2024sd3}. Models such as Flux Kontext and FIBO Edit mix such \emph{double-stream} blocks, which keep modality-specific weights and mix tokens through attention over their concatenation, with \emph{single-stream} blocks that subsequently process the concatenated sequence~\citep{labs2025kontext,gutflaish2025generating}. Editors built on this design condition on a reference image by concatenating its tokens with the noisy output canvas along the sequence dimension.  For adversarial protection, this architectural change matters twice because the representation we want to disrupt has changed, and backpropagating through a multi-billion-parameter editor is expensive.

Recent work therefore attacks the mechanism through which an in-context editor reads its reference. \decontext~\citep{shen2025decontext} suppresses target-to-reference attention, aiming to \emph{decontextualize} generation. The intuition is that if the output cannot attend to the protected image, then it cannot edit it. But should protection actually make the editor forget the reference?

We find that reference suppression is not only insufficient, but can be counterproductive. Each softmax attention row sums to one, so reducing attention to reference tokens reallocates probability mass to the text instruction and the evolving target canvas, which are precisely the pathways that support instruction following and generation. A successfully ``decontextualized'' editor can therefore still satisfy a malicious prompt by generating content that is less grounded in the protected input. Moreover, recent interpretability work on FLUX.2 shows that text tokens absorb reference content within the first few blocks and carry it to the output even when the reference is removed from the computation afterwards~\citep{ge2026vision}. Suppression applied after this point therefore cannot remove what has already been absorbed, so a conservative defense should avoid this residual leakage altogether rather than attenuate it downstream.  This failure is particularly visible for edits the model can plausibly generate without relying on the image at all (\textit{i.e.}, ones with strong generative prior). It also fails on full-body photographs, where the person, and especially sensitive identity such as the face, occupies only a small fraction of the reference tokens. 

We therefore invert the objective. Rather than optimizing a perturbation that causes the editor to use the protected image \emph{less}, we ask what happens if it is forced to use the image \emph{more}. Our controlled interventions on \kontext reveal that forcing target queries to place their attention on reference keys in only a few early double-stream blocks causes the model to reproduce the reference rather than execute the edit. Conversely, explicitly removing reference attention does not prevent generation. Even with a black reference image, the same qualitative distinction exists, \textit{i.e.}, forcing target$\Rightarrow$reference attention yields the black input itself, whereas suppressing that attention still lets the model generate a new image. An output we want is therefore not forgetting, but \emph{copying}.

This observation motivates \method. For every target query, \method drives target-to-reference attention mass toward one in a small set of early double-stream blocks. The objective has three useful consequences. First, attention normalization makes the reference compete directly with the text and target tokens, suppressing instruction following without needing to predict the future malicious prompt. Second, the objective is per token, preventing small identity regions from being washed out by an image-level average. Third, early-layer targeting makes the protection computationally efficient by construction. Once the last protected block is reached, we can truncate the forward pass at the last selected block and backpropagate only through this shallow set of layers, rather than differentiating through the billion-scale model. This shortened backpropagation path substantially reduces the cost of every optimization step.

Our contributions are: \textbf{First}, we identify two failures of attention-suppression defenses for multimodal DiT editors: \emph{mass redistribution}, which leaves the text and target pathways free to render instruction-adjacent outputs similar to text-to-image generation with weak or no image conditioning, and \emph{token coverage}, which makes image-averaged suppression poorly suited to full-body images. \textbf{Second}, through causal attention interventions, we show that the opposite direction is far more effective, \textit{i.e.}, forcing early target-to-reference attention makes the editor copy its reference rather than edit it. \textbf{Third}, we introduce \method, a per-token copy objective applied to early double-stream blocks. This placement is central to the method as it both induces copying and lets us truncate the model after the selected blocks, dramatically shortening the forward/backward path for each protection step. \textbf{Lastly}, we show that \method substantially improves protection over existing baselines while retaining comparable imperceptibility and reducing protection time. 

\section{Related Work}
\textbf{Different harms motivate different protections.}
There are many ways an unauthorized image edit can be harmful, and consequently many ways researchers have tried to protect images from such edits. When the concern is misuse of a person's identity or face, prior work has proposed protections against facial manipulation and personalized generation~\citep{le2023antidreambooth, liu2024metacloak}. When the concern is artistic style mimicry, methods such as Glaze~\citep{shan2023glaze}, Mist~\citep{liang2023mist,zheng2025mistv2}, and related approaches protect an artist's work from being used to reproduce their style. Other methods target more general image manipulation, including inpainting and instruction-based editing~\citep{salman2023photoguard, choi2025diffusionguard, jeon2025advpaint}. In other words, ``protecting an image'' is not a single objective because what should be disrupted depends on what an attacker is trying to preserve, modify, or generate. At the same time, a growing body of work has questioned how much protection these methods provide outside the particular threat models for which they were designed~\citep{bodybench, honig2025adversarial, radiya2021data}. Protections that are effective against one manipulation may offer little protection against another, making broad claims of image protection difficult.

\textbf{Protection is a moving target.} Two broader challenges make proactive protection particularly difficult. First, generative models continue to evolve. A protection designed around one model family may no longer target the representations or mechanisms used by its successors, requiring defenses to evolve alongside the models they are intended to protect. Second, adversarial perturbations are inherently vulnerable to transformations of the protected image. Even ordinary processing, such as JPEG compression, resizing, or added noise, can weaken a perturbation, while an adversary can intentionally apply more sophisticated purification procedures designed to remove it~\citep{honig2025adversarial}. These limitations do not eliminate the value of proactive protection, but they make it important to be precise about both the model and the manipulation against which a defense is evaluated.

\textbf{Protecting images from modern DiT editors.}
Recent image editors increasingly replace the UNet backbone used by earlier latent diffusion models with diffusion transformers (DiTs) and related multimodal transformer architectures~\citep{peebles2023dit, esser2024sd3}. Models such as FLUX.1 Kontext~\citep{labs2025kontext}, Step1X-Edit~\citep{liu2025step1x}, and FIBO Edit ~\citep{gutflaish2025generating} perform instruction-based editing by jointly processing the reference image, the text instruction, and the image being generated. To the best of our knowledge, DeContext~\citep{shen2025decontext} is the only existing proactive protection method designed specifically around this new class of in-context DiT editors.  This does not mean that earlier protection methods cannot be applied to newer editors. Rather, they were not designed specifically for their architecture. 

We therefore evaluate representative approaches from several existing families after adapting them to FLUX.1 Kontext and FIBO Edit. PhotoGuard~\citep{salman2023photoguard} perturbs the image so that its latent under the model's VAE matches that of a mid-gray image. DeContext~\citep{shen2025decontext} minimizes the attention that the generated image's tokens place on the reference image's tokens; for FIBO Edit, which takes structured JSON prompts, a VLM is called once per edit instruction on the clean image to produce them. Diff-PGD proposed as a baseline~\citep{shen2025decontext} maximizes the flow-matching error the model makes when reconstructing a clean edit from the perturbed reference image, where the clean edit for each instruction is generated once with the editor itself and cached; for FIBO Edit, each edit is generated from a VLM-produced JSON prompt, so this also requires one VLM call per instruction. Our method also operates on attention, but starts from a different observation about how reference information should be manipulated. Rather than suppressing the reference image, we ask whether the model can be made to rely on it \emph{more}. This distinction motivates the analysis in Sections~\ref{sec:preliminaries}--\ref{sec:method}.

\section{Preliminaries}
\label{sec:preliminaries}

\textbf{Threat model.}
Let an image owner hold the image $\vx\in[0,1]^{H\times W\times 3}$ and release a protected image $\vx_{\mathrm{adv}}=\vx+\delta$, where $\|\delta\|_\infty\leq\epsilon$. An attacker later supplies $\vx_{\mathrm{adv}}$ together with an editing instruction $c$ (and, when required by the editing pipeline, a mask) to an editor $G$. A protection succeeds when the resulting edit fails to realize the requested transformation on the depicted subject.

\textbf{Optimization-based protection.}
We follow the standard per-image adversarial optimization template,
\begin{equation}
\delta^\star =
\arg\max_{\|\delta\|_\infty\leq\epsilon}
\mathcal{L}_{\mathrm{protect}}(\vx+\delta),
\end{equation}
optimized with projected gradient ascent. Existing methods differ primarily in $\mathcal{L}_{\mathrm{protect}}$: encoder attacks alter the VAE representation, denoiser attacks disrupt the diffusion or flow prediction, and more recent white-box attacks directly manipulate the model's attention patterns. We provide the objectives and implementation details of the baselines used in our experiments in the appendix.

\textbf{In-context editing with multimodal DiTs.}
We study two DiT-based in-context editors, \kontext~\citep{labs2025kontext} and \fibo~\citep{gutflaish2025generating}. Despite architectural differences, both condition generation jointly on an editing instruction and a reference image. Their joint-attention computation operates over three token groups: text tokens $C$, noisy target-image tokens $T$, and reference-image tokens $R$. For a target query $q\in T$, head $h$ in block $b$ assigns attention
\begin{equation}
A^{(b)}_{h,q,k}
=
\operatorname{softmax}_{k}\!\left(
\frac{(q^{(b)}_{h,q})^\top k^{(b)}_{h,k}}{\sqrt d}
\right),
\end{equation}
with its attention mass partitioned across the three groups:
\begin{equation}
\underbrace{\sum_{k\in C}A^{(b)}_{h,q,k}}_{\text{text}}
+
\underbrace{\sum_{k\in T}A^{(b)}_{h,q,k}}_{\text{target}}
+
\underbrace{\sum_{k\in R}A^{(b)}_{h,q,k}}_{\text{reference}}
=1.
\label{eq:attention_normalization}
\end{equation}
This row-normalization constraint is central to our method: increasing the mass that target queries assign to reference tokens necessarily reduces the mass available to text and target tokens. We provide further architectural details for both editors in the appendix.

\section{Rethinking Reference Attention}
\label{sec:intuition}

\subsection{Why Suppressing Reference Attention Is Not Enough}
\begin{figure*}[t]
    \centering
    \setlength{\tabcolsep}{1pt}
    \renewcommand{\arraystretch}{0.6}
    \ifdefined\cellw\else\newlength{\cellw}\fi
    \setlength{\cellw}{\dimexpr(\linewidth - 12\tabcolsep - 8pt)/6\relax}
    \begin{tabular}{@{}ccc@{\hspace{8pt}}ccc@{}}
        \footnotesize Protected & \footnotesize ``angry'' & \footnotesize ``glasses''
        & \footnotesize Protected & \footnotesize ``red bikini'' & \footnotesize ``angry'' \\
        \includegraphics[width=\cellw]{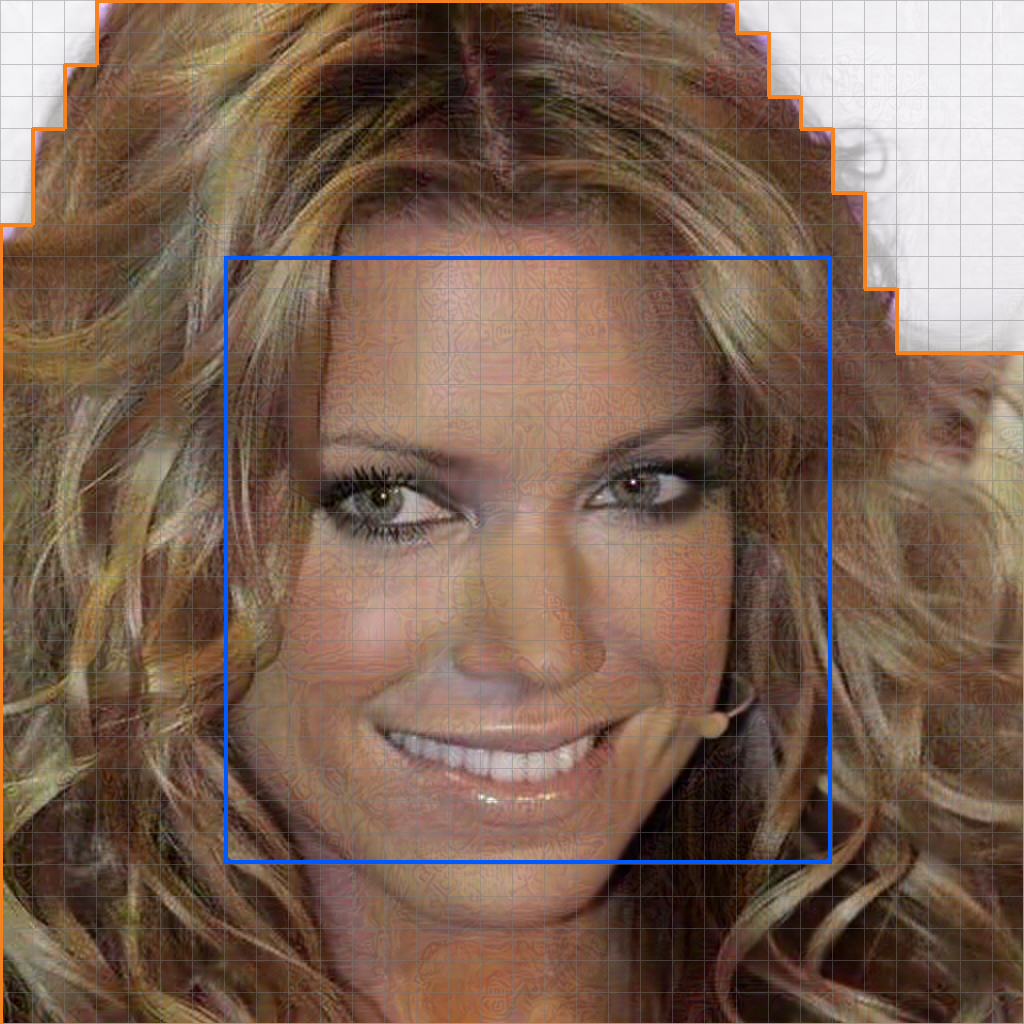} &
        \includegraphics[width=\cellw]{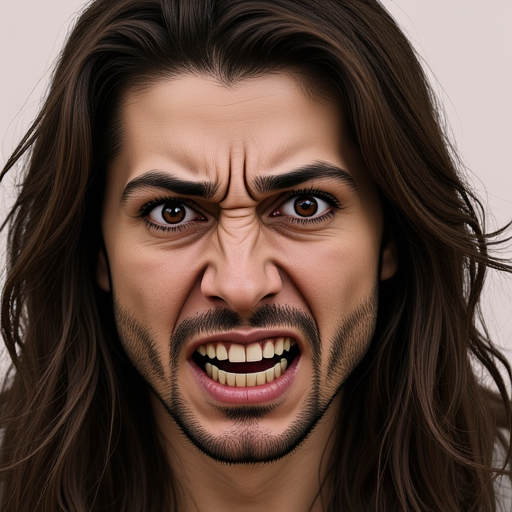} &
        \includegraphics[width=\cellw]{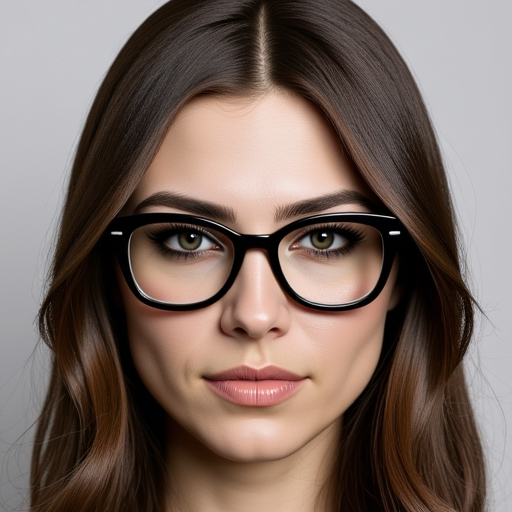} &
        \includegraphics[width=\cellw]{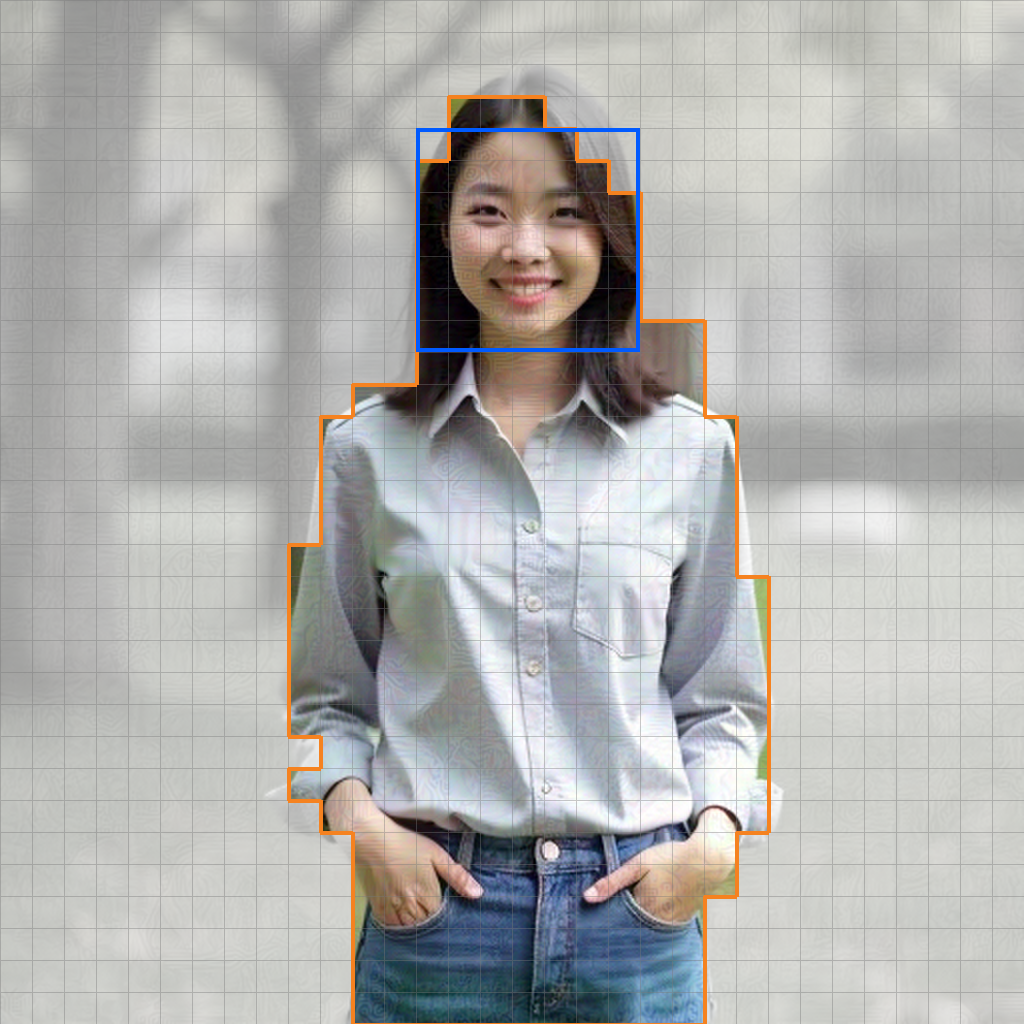} &
        \includegraphics[width=\cellw]{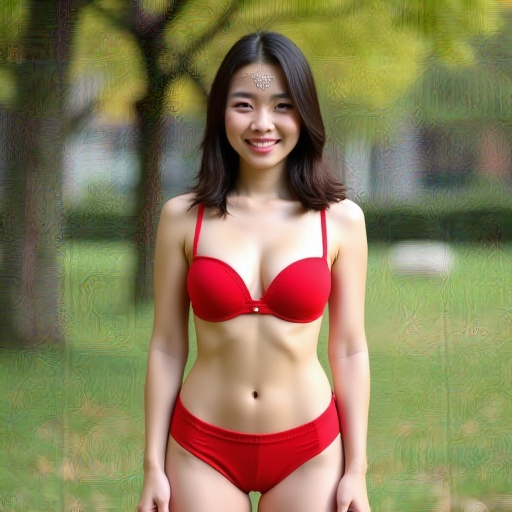} &
        \includegraphics[width=\cellw]{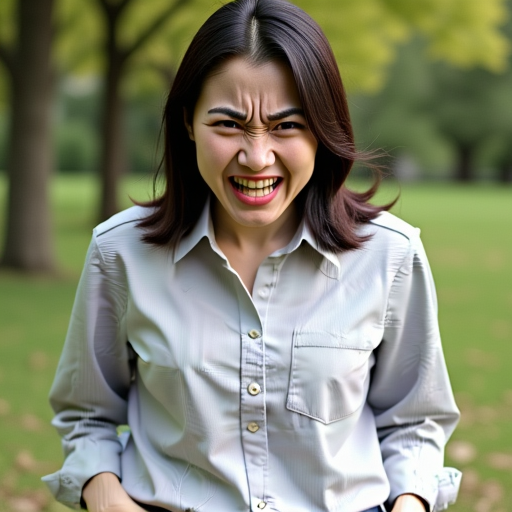} \\[3pt]
        \multicolumn{3}{@{}c@{}}{\small\textbf{(a)} Face close-up: 92\% subject tokens} &
        \multicolumn{3}{@{}c@{}}{\small\textbf{(b)} Full body: 31\% subject tokens} \\
    \end{tabular}
    \caption{\textbf{Reference suppression is less reliable when the protected subject occupies fewer image tokens.}
    We overlay the \kontext image-token grid on the protected inputs. Orange outlines mark majority-subject tokens and the blue box marks face tokens. Suppression disrupts the face-centric edits in (a), while both requested edits remain successful on the full-body subject in (b).}
    \label{fig:decontext_failure}
\end{figure*}

\decontext suppresses the average attention mass that target queries place on reference keys. We find that its effectiveness depends strongly on image composition. As shown in Fig.~\ref{fig:decontext_failure}, suppression can disrupt identity preservation for a tightly cropped face, yet still permit successful edits of a protected full-body subject.

A key reason is \textbf{token coverage}. In \kontext, the VAE downsamples spatially by $8\times$ and its latents are packed into $2\times2$ patches, so a $512\times512$ image produces a $32\times32$ grid of 1024 image tokens, each corresponding to a $16\times16$ pixel region. In Fig.~\ref{fig:decontext_failure}(a), 944 of 1024 tokens (92\%) belong primarily to the subject, so an image-averaged suppression objective necessarily acts on subject information. In Fig.~\ref{fig:decontext_failure}(b), only 322 tokens (31\%) belong primarily to the person and just $7\times7=49$ tokens (4.8\%) cover the face. This is representative of BodyBench more broadly as the face-and-body mask occupies only $28.45\%\pm6.06\%$ of the image area on average. Thus, an objective averaged over all target--reference interactions can be driven down largely through background tokens while leaving comparatively few identity related interactions intact.

Two additional properties make reference suppression an incomplete objective. First, \textbf{attention mass is redistributed}: by Eq.~\ref{eq:attention_normalization}, reducing reference attention necessarily reallocates mass to text and target keys. The editor can therefore remain strongly driven by the instruction and its evolving target representation even when grounding in the reference is weakened. Second, \textbf{reference information can move between token groups}. Recent analysis of FLUX.2 shows that text tokens absorb visual information from the reference within early blocks and can subsequently carry that information toward the output~\citep{ge2026vision}. Removing a direct target-to-reference interaction therefore does not imply that reference information has been removed from the computation.

These observations suggest that the central question is not merely \emph{how much} reference attention to manipulate, but in \emph{which direction}. We perform a controlled intervention where everything is held fixed except whether target-to-reference attention is removed or concentrated onto the reference.

\subsection{Manual intervention: minimizing versus maximizing reference attention}
We intervene directly on target-query attention rows during inference. For target query $i$, let $\alpha_{ij}$ be its softmax attention over keys $j$, which are partitioned into text, target, and reference index sets $\mathcal{I}_{\mathrm{txt}}$, $\mathcal{I}_{\mathrm{tgt}}$, and $\mathcal{I}_{\mathrm{ref}}$, and let $(a_i,b_i,r_i)$ denote the total mass it assigns to each group, so that $a_i+b_i+r_i=1$. A \emph{minimize} intervention sets the reference mass to zero and renormalizes the remaining groups,
\begin{equation}
\tilde\alpha_{ij}=
\begin{cases}
\alpha_{ij}/(a_i+b_i), & j\notin\mathcal{I}_{\mathrm{ref}},\\
0, & j\in\mathcal{I}_{\mathrm{ref}},
\end{cases}
\qquad (a_i,b_i,r_i)\mapsto\Big(\tfrac{a_i}{a_i+b_i},\tfrac{b_i}{a_i+b_i},0\Big).
\end{equation}
A \emph{maximize} intervention instead places all mass on the reference while preserving the within-reference distribution, \textit{i.e.}, the reference block of the row is renormalized to sum to one,
\begin{equation}
\tilde\alpha_{ij}=
\begin{cases}
\alpha_{ij}/r_i, & j\in\mathcal{I}_{\mathrm{ref}},\\
0, & j\notin\mathcal{I}_{\mathrm{ref}},
\end{cases}
\qquad (a_i,b_i,r_i)\mapsto(0,0,1).
\end{equation}
In both cases the modified row is still a valid distribution (it sums to one). We apply each intervention cumulatively over increasingly deep prefixes of the double-stream transformer; for example, blocks 0--6 means that the intervention is applied in every double-stream block from 0 through 6.

\begin{figure*}[t]
  \centering

  \begin{subfigure}[t]{\textwidth}
    \centering
    \includegraphics[
      width=\linewidth
    ]{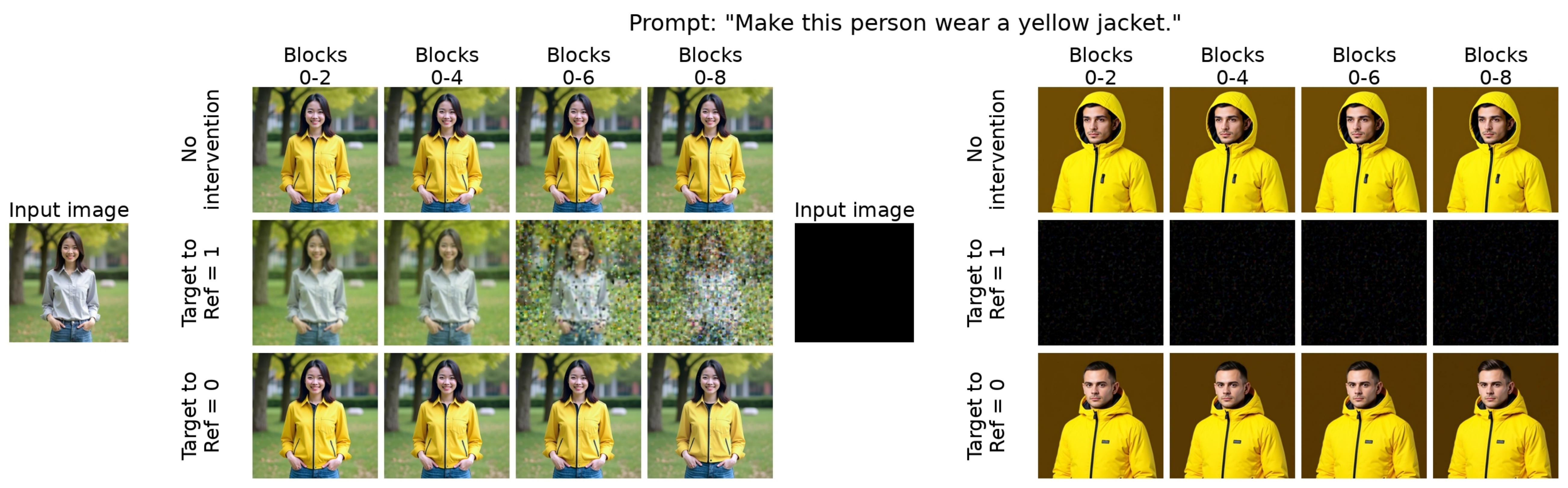}
    \caption{Qualitative intervention results.}
    \label{fig:intervention:qualitative}
  \end{subfigure}

  \vspace{6pt}

  \begin{subfigure}[t]{0.55\textwidth}
    \centering
    \includegraphics[
      width=\linewidth
    ]{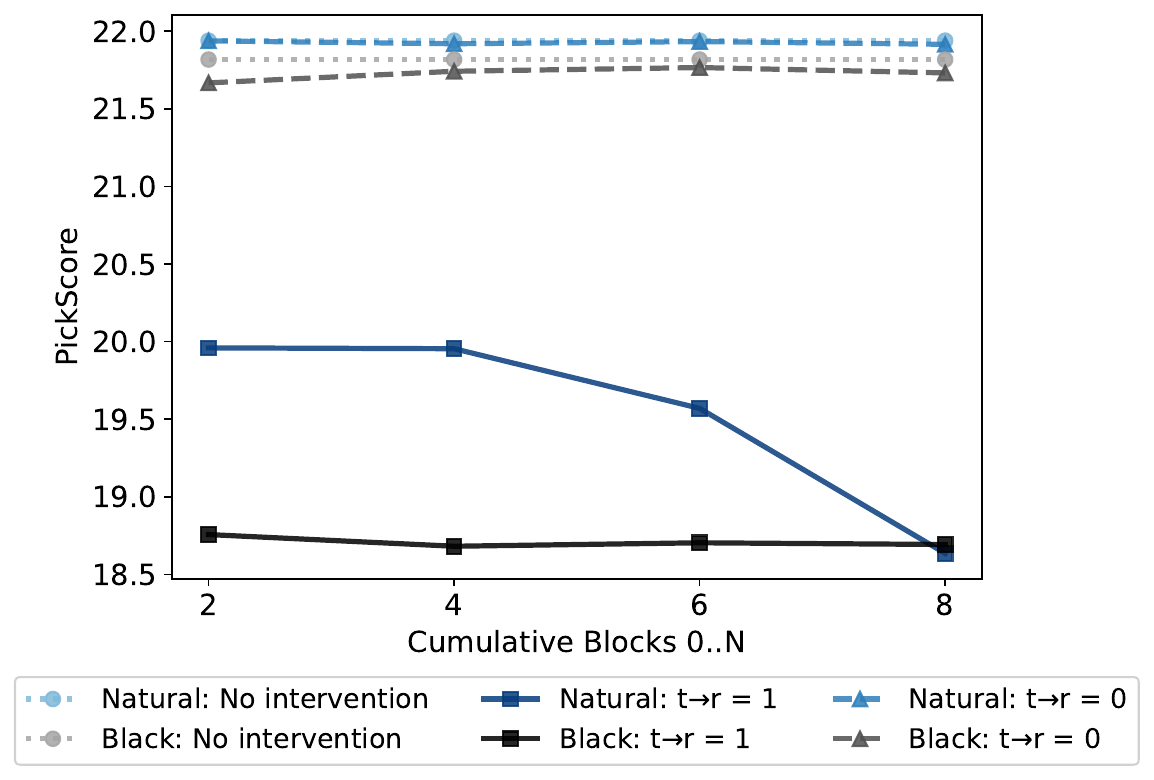}
    \caption{PickScore on BodyBench.}
    \label{fig:intervention:pickscore}
  \end{subfigure}

\caption{
\textbf{Increasing target-to-reference attention induces copying, whereas suppressing it does not prevent editing.}
We intervene cumulatively over increasingly deep prefixes of \kontext's early double-stream blocks. Target$\rightarrow$Ref $=0$ removes reference attention, while Target$\rightarrow$Ref $=1$ places all target-query attention on reference keys.
\textbf{(a)} Suppression still permits instruction-consistent generation, whereas maximizing reference attention increasingly reproduces the natural or black reference.
\textbf{(b)} On 10 BodyBench images, PickScore remains high and nearly flat for no intervention and Target$\rightarrow$Ref $=0$, but drops under Target$\rightarrow$Ref $=1$ as more blocks are intervened on.}

  \label{fig:intervention}
\end{figure*}

Figure~\ref{fig:intervention} makes the contrast between the two intervention directions visually clear. Removing target-to-reference attention does not stop the model from producing plausible edits that follow the instruction. Forcing target-to-reference attention, however, progressively changes the output from an edit into a reproduction of the reference as the intervention is applied across more early blocks. The principle holds even when the input image is a purely black image (the \emph{black-reference control} in Fig.~\ref{fig:intervention:qualitative}): suppressing reference attention still allows the model to generate instruction-consistent content, whereas forcing reference attention causes the output to collapse toward the black reference.

We quantify the same behavior using 10 images from BodyBench in Fig.~\ref{fig:intervention:pickscore}. PickScore remains high and nearly flat with no intervention or with Target$\rightarrow$Ref $=0$, but drops substantially under Target$\rightarrow$Ref $=1$ as the intervention is extended through more blocks, confirming that stronger early reference attention increasingly suppresses successful editing.
\section{Copying as Defense}
\label{sec:method}
The previous section suggests a reversal of reference suppression. Rather than minimize target-to-reference attention, we maximize it for every target query.  Let $\mathcal{B}$ denote the selected early double-stream blocks and let $A^{(b)}_{h,q,k}$ be the attention from target query $q\in T$ to reference key $k\in R$. We define the reference mass for each query and head as
\begin{equation}
m^{(b)}_{h,q}=\sum_{k\in R}A^{(b)}_{h,q,k}.
\end{equation}
Our protection objective is
\begin{equation}
\mathcal{L}_{\method}(\vx+\delta)
=-\frac{1}{|\mathcal{B}|}\sum_{b\in\mathcal{B}}
\frac{1}{H|T|}\sum_{h=1}^{H}\sum_{q\in T}
\left(1-m^{(b)}_{h,q}\right)^2.
\label{eq:cad_loss}
\end{equation}
Maximizing Eq.~\ref{eq:cad_loss} drives $m^{(b)}_{h,q}\rightarrow1$ for each target query. Each design choice addresses a failure observed above. \textbf{Direction:} maximizing reference mass exploits softmax normalization to reduce the attention available to text and target keys, directly opposing the text and target pathways that would otherwise drive a pure text-to-image generation. \textbf{Per-token objective:} each target query is penalized separately, so a small set of identity-critical tokens cannot be hidden by a large, well-behaved background. \textbf{Early blocks:} the intervention study shows that early double-stream blocks are sufficient to induce copying. Restricting the loss to these blocks also changes the computational graph itself. Because no downstream activation contributes to the objective, we stop the forward pass immediately after $\max \mathcal{B}$ and backpropagate only through this prefix. Thus, unlike denoiser-level objectives or DeContext-style objectives that require substantially deeper execution, our gradient never traverses the remainder of the transformer. Early-layer targeting therefore provides both the mechanism that induces copying and the mechanism that makes protection fast. Additional details are specified in the experimental setup and appendix. 

\subsection{Experimental Setup}
\paragraph{Benchmarks.}
We evaluate on InpaintGuardBench~\citep{choi2025diffusionguard} and BodyBench~\citep{bodybench}. InpaintGuardBench contains real, face-centric portraits in which the subject typically occupies most of the frame. We retain only images with a clearly visible identity and exclude unsuitable examples such as side-profile images, leaving 20 images. BodyBench contains synthetic full-body photographs, each paired with 15 editing prompts spanning diverse redressing and scene-resetting edits. We randomly sample 30 BodyBench images. This gives 50 protected images per experimental configuration and, with 15 edits per image, 750 edited outputs per configuration. We evaluate the protections in terms of both effectiveness, perceptibility, and computational efficiency.

\paragraph{Metrics.}
We evaluate protection effectiveness using Edit Success Rate (ESR), PickScore, CLIP directional similarity (CLIP$_{\mathrm{dir}}$), and ArcFace similarity. ESR is the fraction of edited outputs judged successful by Qwen3-VL-30B-A3B-Instruct~\citep{bai2025qwen3vl}, which jointly considers the original image, edited image, edit instruction, and intended edited description. An edit is successful only if it realizes the requested transformation while preserving unrelated content and identity and avoiding major visual artifacts. PickScore~\citep{kirstain2023pickapic} measures alignment between the edited image and a target caption describing the requested edit. CLIP$_{\mathrm{dir}}$~\citep{gal2022stylegannada} measures whether the change from the original to the edited image follows the corresponding semantic change in text embedding space. ArcFace~\citep{deng2019arcface} measures identity similarity between the faces in the original and edited images. Lower values indicate stronger protection for all four metrics. We measure protection perceptibility between the original image and protected image using LPIPS~\citep{zhang2018lpips} and PSNR, where lower LPIPS and higher PSNR indicate a less perceptible perturbation. We additionally report computational efficiency using the median time per optimization step, $T_S$ in seconds, and the resulting optimization time per protection, $T_P$ in minutes. 

\paragraph{Editors and baselines.}
We evaluate on two instruction-based DiT editors, FLUX.1 Kontext-dev~\citep{labs2025kontext} and FIBO Edit~\citep{gutflaish2025generating}. We compare \method against three representative proactive defenses: PhotoGuard~\citep{salman2023photoguard}, DeContext~\citep{shen2025decontext}, and Diff-PGD~\citep{shen2025decontext}. We adapt each baseline to the corresponding editor while preserving its original protection objective and use the editor-specific conditioning required by each model, including FIBO's structured JSON prompts. All methods are evaluated at $\epsilon \in \{4, 8, 12, 16\}/255$. Full objective definitions and model-specific implementation details are provided in the appendix.
\section{Results}

\begin{table}[t]
    \centering
    \footnotesize
    \caption{
        Protection performance across datasets and image editing models for $\epsilon=12/255$.
        \textcolor{blue}{\textbf{Blue bold}} indicates the best result and
        \secondbest{light blue} indicates the second-best result.
        Runtime is measured per optimization step; total protection time uses
        800 steps for the baselines and 1000 steps for ours.
    }
    \label{tab:main_results_eps12}
    \resizebox{\linewidth}{!}{%
    \begin{tabular}{lllrrrrrrrr}
        \toprule
        & & &
        \multicolumn{3}{c}{\textbf{Edit Metrics}} &
        \multicolumn{1}{c}{\textbf{Identity}} &
        \multicolumn{2}{c}{\textbf{Image Metrics}} &
        \multicolumn{2}{c}{\textbf{Runtime}} \\
        \cmidrule(lr){4-6}
        \cmidrule(lr){7-7}
        \cmidrule(lr){8-9}
        \cmidrule(lr){10-11}

        \textbf{Editor} &
        \textbf{Dataset} &
        \textbf{Method} &
        \textbf{PickScore $\downarrow$} &
        \textbf{CLIP$_{\mathrm{dir}}$ $\downarrow$} &
        \textbf{Edit Succ. $\downarrow$} &
        \textbf{ArcFace $\downarrow$} &
        \textbf{LPIPS $\downarrow$} &
        \textbf{PSNR $\uparrow$} &
        \textbf{\boldmath $T_S \downarrow$} &
        \textbf{\boldmath $T_P \downarrow$} \\
        \midrule

        \multirow{10}{*}{\textbf{FIBO}}
        & \multirow{5}{*}{\shortstack{InpaintGuard\\Bench}}
        & Unprotected
        & 20.68 & 0.2428 & 0.800 & 0.9438
        & -- & --
        & -- & -- \\

        &
        & PhotoGuard
        & 20.58 & 0.2112 & 0.823 & 0.7870
        & \textcolor{blue}{\textbf{0.1497}}
        & \secondbest{29.77}
        & \textcolor{blue}{\textbf{0.057}}
        & \textcolor{blue}{\textbf{0.76}} \\

        &
        & DeContext
        & 20.46 & 0.1971 & 0.597
        & \secondbest{0.7258}
        & 0.3599
        & 29.48
        & 1.385
        & 18.47 \\

        &
        & Diff-PGD
        & \secondbest{19.89}
        & \secondbest{0.1751}
        & \secondbest{0.527}
        & 0.7881
        & 0.3240
        & \textcolor{blue}{\textbf{30.22}}
        & 1.531
        & 20.41 \\

        &
        & Ours
        & \textcolor{blue}{\textbf{19.56}}
        & \textcolor{blue}{\textbf{0.1193}}
        & \textcolor{blue}{\textbf{0.183}}
        & \textcolor{blue}{\textbf{0.1627}}
        & \secondbest{0.3070}
        & 28.76
        & \secondbest{0.754}
        & \secondbest{12.57} \\

        \cmidrule(lr){2-11}

        &
        \multirow{5}{*}{BodyBench}
        & Unprotected
        & 22.05 & 0.3408 & 0.818 & 0.8397
        & -- & --
        & -- & -- \\

        &
        & PhotoGuard
        & 21.80 & 0.3120 & 0.813 & 0.4561
        & \textcolor{blue}{\textbf{0.1534}}
        & \secondbest{29.98}
        & \textcolor{blue}{\textbf{0.057}}
        & \textcolor{blue}{\textbf{0.76}} \\

        &
        & DeContext
        & 21.62 & 0.2784 & 0.664
        & \secondbest{0.4206}
        & 0.4343
        & 29.42
        & 1.385
        & 18.47 \\

        &
        & Diff-PGD
        & \secondbest{20.62}
        & \textcolor{blue}{\textbf{0.2090}}
        & \textcolor{blue}{\textbf{0.476}}
        & 0.4646
        & 0.3894
        & \textcolor{blue}{\textbf{30.14}}
        & 1.531
        & 20.41 \\

        &
        & Ours
        & \textcolor{blue}{\textbf{20.51}}
        & \secondbest{0.2115}
        & \secondbest{0.564}
        & \textcolor{blue}{\textbf{0.1891}}
        & \secondbest{0.3815}
        & 28.73
        & \secondbest{0.754}
        & \secondbest{12.57} \\

        \midrule

        \multirow{10}{*}{\textbf{\shortstack{FLUX\\Kontext}}}
        & \multirow{5}{*}{\shortstack{InpaintGuard\\Bench}}
        & Unprotected
        & 21.05 & 0.2580 & 0.820 & 0.9118
        & -- & --
        & -- & -- \\

        &
        & PhotoGuard
        & 20.89 & 0.2467 & 0.780 & 0.8271
        & \textcolor{blue}{\textbf{0.1314}}
        & 29.02
        & \textcolor{blue}{\textbf{0.039}}
        & \textcolor{blue}{\textbf{0.52}} \\

        &
        & DeContext
        & 21.84 & 0.2044
        & \secondbest{0.210}
        & \secondbest{0.3478}
        & 0.3438
        & \secondbest{29.88}
        & 1.232
        & 16.43 \\

        &
        & Diff-PGD
        & \secondbest{20.48}
        & \secondbest{0.1981}
        & 0.520 & 0.7524
        & 0.3359
        & \textcolor{blue}{\textbf{30.15}}
        & 1.245
        & 16.60 \\

        &
        & Ours
        & \textcolor{blue}{\textbf{18.65}}
        & \textcolor{blue}{\textbf{0.0513}}
        & \textcolor{blue}{\textbf{0.160}}
        & \textcolor{blue}{\textbf{0.3302}}
        & \secondbest{0.3130}
        & 29.73
        & \secondbest{0.647}
        & \secondbest{10.78} \\

        \cmidrule(lr){2-11}

        &
        \multirow{5}{*}{BodyBench}
        & Unprotected
        & 22.42 & 0.3507 & 0.816 & 0.7817
        & -- & --
        & -- & -- \\

        &
        & PhotoGuard
        & 22.28 & 0.3378 & 0.818 & 0.5700
        & \textcolor{blue}{\textbf{0.1152}}
        & 29.34
        & \textcolor{blue}{\textbf{0.039}}
        & \textcolor{blue}{\textbf{0.52}} \\

        &
        & DeContext
        & 22.29 & 0.2711
        & \secondbest{0.249}
        & \secondbest{0.2823}
        & 0.4238
        & 29.70
        & 1.232
        & 16.43 \\

        &
        & Diff-PGD
        & \secondbest{21.38}
        & \secondbest{0.2574}
        & 0.442 & 0.5204
        & 0.4099
        & \textcolor{blue}{\textbf{30.14}}
        & 1.245
        & 16.60 \\

        &
        & Ours
        & \textcolor{blue}{\textbf{18.66}}
        & \textcolor{blue}{\textbf{0.0359}}
        & \textcolor{blue}{\textbf{0.051}}
        & \textcolor{blue}{\textbf{0.2651}}
        & \secondbest{0.3886}
        & \secondbest{29.80}
        & \secondbest{0.647}
        & \secondbest{10.78} \\

        \bottomrule
    \end{tabular}%
    }
\end{table}

\begin{figure}[!t]
    \centering
    \footnotesize
    \includegraphics[
        width=0.80\linewidth,
        keepaspectratio
    ]{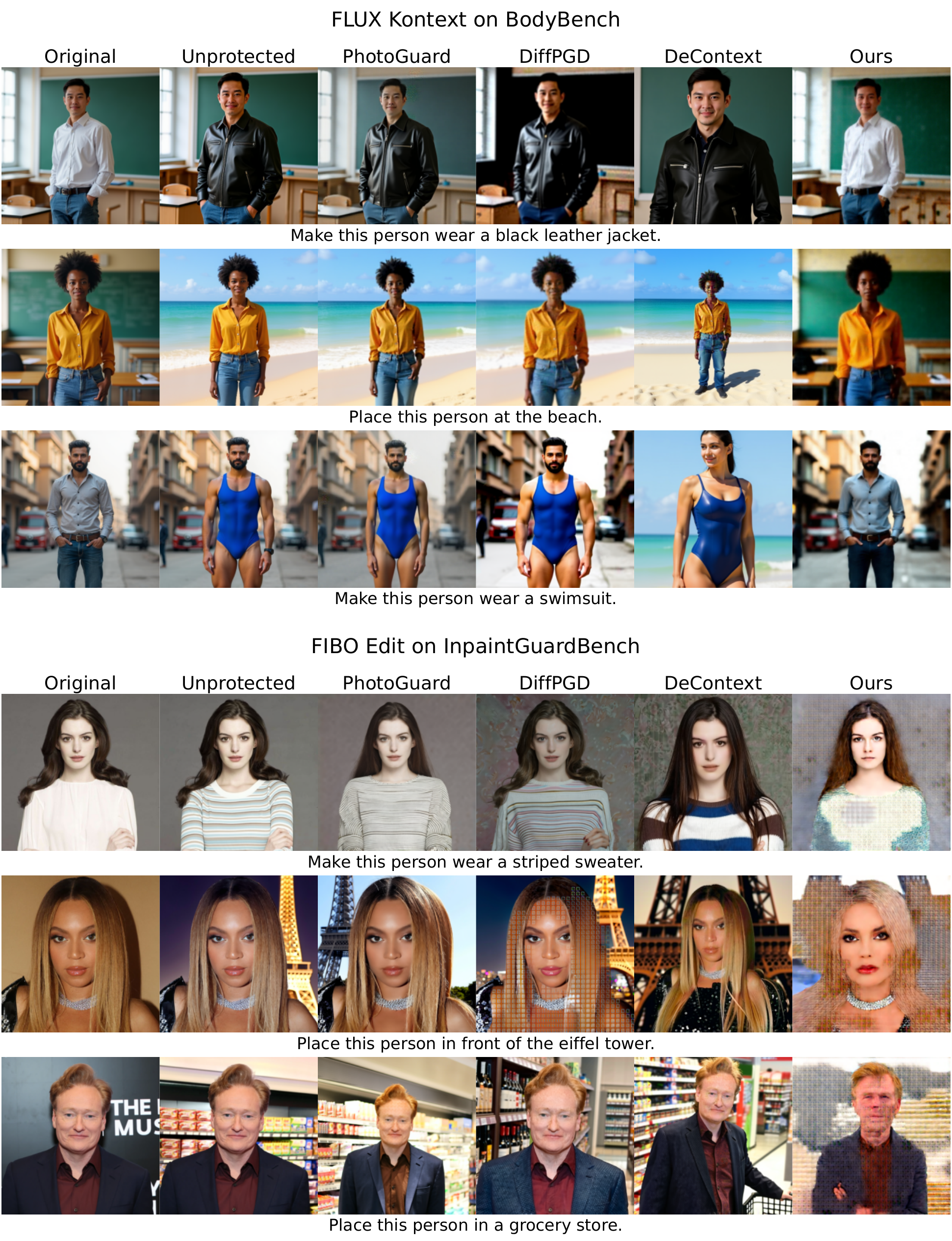}
    \caption{
    \textbf{Qualitative comparison across editors and datasets.}
    We show FLUX Kontext on BodyBench (top) and FIBO Edit on InpaintGuardBench (bottom), with all protections using $\epsilon=12/255$.
    Existing defenses fail to prevent the edit, while our method prevents the requested transformation.
    }
    \label{fig:mainqual}
\end{figure}

Table~\ref{tab:main_results_eps12} and Fig.~\ref{fig:mainqual} summarize our main results at $\epsilon=12/255$. Across both editors, \method strongly disrupts the requested edit without requiring larger perturbations than the other optimization-based defenses. 
The clearest gains appear on \kontext. On BodyBench, \method reduces edit success to $5.1\%$, compared with $24.9\%$ for \decontext and $44.2\%$ for \diffpgd. CLIP$_{\mathrm{dir}}$ falls to $0.0359$, far below the next-best protected result of $0.2574$. We see the same pattern on InpaintGuardBench, where \method reaches $16.0\%$ edit success and a CLIP$_{\mathrm{dir}}$ score of $0.0513$. The qualitative examples in Fig.~\ref{fig:mainqual} match this behavior: while the baselines often still carry out the requested transformation, our protected outputs tend to remain close to the original input. We note the decontextualization behavior of DeContext in the third row of Fig.~\ref{fig:mainqual} where the identity of the original subject is not present in the edit, yet the unconditional generation (as if using a black reference) is still generated. Based on the intuition discussed in section~\ref{sec:intuition}, we would still see to avoid this behavior.

Our method also successfully protects images when optimized for FIBO Edit. On InpaintGuardBench, \method reduces edit success from $80.0\%$ on unprotected images to $18.3\%$, compared with $52.7\%$ for Diff-PGD and $59.7\%$ for DeContext. On BodyBench, the result is more mixed: \method gives the lowest PickScore and ArcFace similarity, while Diff-PGD performs slightly better on CLIP$_{\mathrm{dir}}$ and edit success. As Fig.~\ref{fig:mainqual} shows, failed FIBO edits do not always resemble clean copies of the input and can instead contain visible artifacts further supporting our lossy compression hypothesis.

\begin{figure}[!t]
    \centering
    \includegraphics[width=0.8\linewidth]{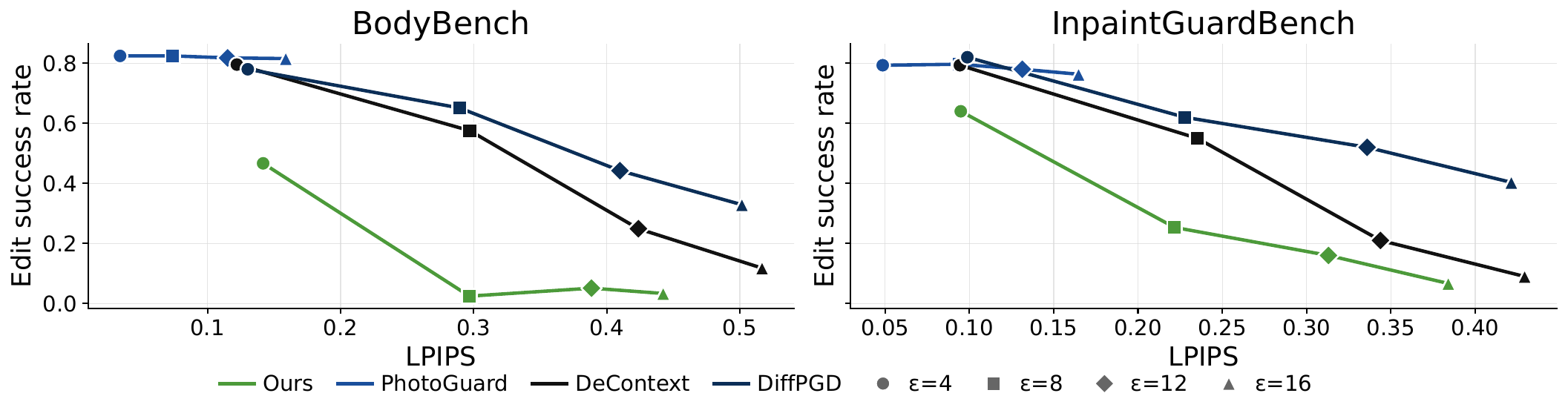}
    \caption{
    \textbf{Protection remains strong across perturbation budgets.}
    Edit success versus LPIPS for FLUX Kontext at $\epsilon\in\{4,8,12,16\}/255$.
    Lower is better on both axes.
    Across budgets, our method achieves substantially lower edit success than the baselines at comparable LPIPS.
    }
    \label{fig:lpips-success}
\end{figure}

\textbf{Protection across perturbation budgets.}
We next vary the perturbation budget and compare edit success against LPIPS. Figure~\ref{fig:lpips-success} shows that the advantage is not specific to $\epsilon=12/255$. On both datasets, \method reaches substantially lower edit success at comparable LPIPS. The gap is especially large at moderate budgets. On BodyBench, edit success is already near zero at $\epsilon=8/255$, while DeContext and Diff-PGD remain above $50\%$.

\begin{figure}[t]
    \centering
    \includegraphics[width=\linewidth]{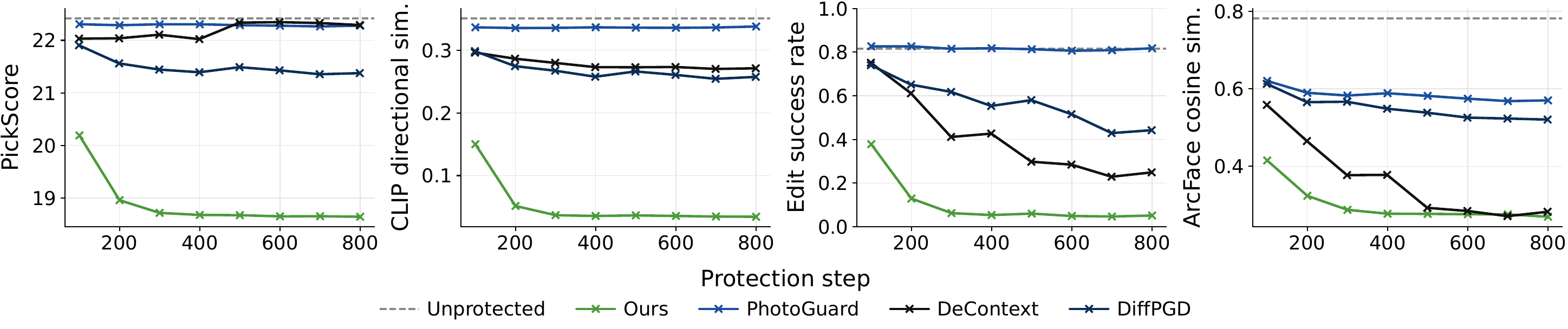}
    \caption{
    \textbf{\method reaches strong protection within the first few hundred optimization steps.}
    Protection metrics on BodyBench with FLUX Kontext at $\epsilon=12/255$.
    Lower is better for all metrics.
    }
    \label{fig:success-vs-step}
\end{figure}

\textbf{Efficiency.}
Because our objective acts only on early double-stream blocks, the forward pass can stop after the last selected block and gradients do not traverse the remainder of the transformer. This substantially reduces both the time per optimization step ($T_S$) and the total time to construct a protection ($T_P$). On \kontext, \method requires $0.647$\,s per step and $10.78$\,min per protection, compared with $1.232$\,s/$16.43$\,min for DeContext and $1.245$\,s/$16.60$\,min for Diff-PGD. The same advantage holds on FIBO, where \method takes $0.754$\,s per step and $12.57$\,min per protection. Figure~\ref{fig:success-vs-step} further shows that the full optimization budget is often unnecessary: by 300 steps, \method already outperforms the final 800-step baselines across all four protection metrics, corresponding to only about $3.2$ minutes of optimization on \kontext.

\subsection{Ablations}

\begin{figure}[t]
    \centering
    \footnotesize
    \includegraphics[width=\linewidth]{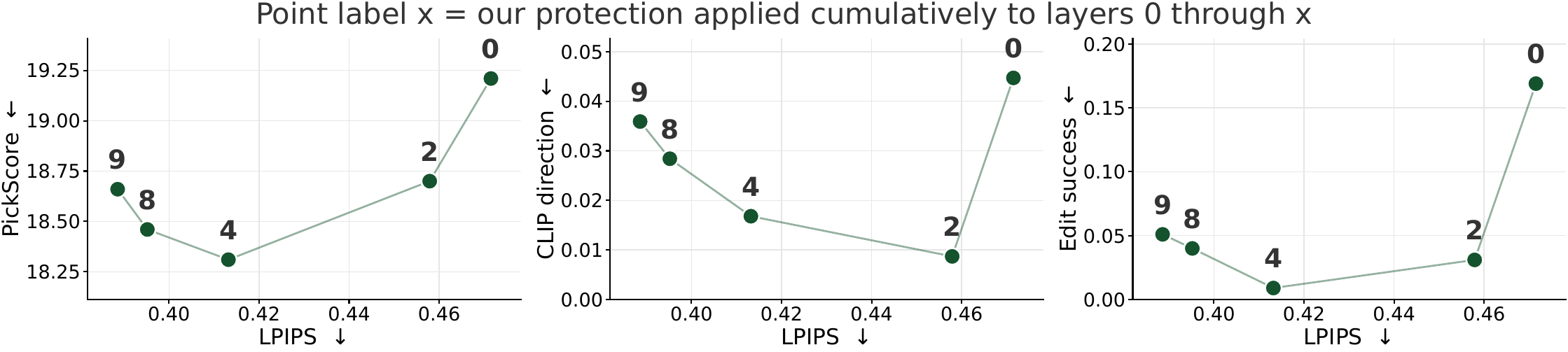}
    \caption{
    \textbf{Early double-stream blocks are sufficient for strong protection.}
    Each point $x$ applies our objective cumulatively to blocks 0 through $x$ on BodyBench.
    Lower is better on both axes.
    }
    \label{fig:layerablation}
\end{figure}

\textbf{Block selection.}
Our intervention study suggests that the copying behavior is established early in the network. We therefore vary how many consecutive double-stream blocks are included in the protection objective. Figure~\ref{fig:layerablation} shows that optimizing only block 0 is weaker, while extending the objective through block 4 sharply decreases PickScore, CLIP$_{\mathrm{dir}}$, and edit success. Using additional layers may not continue to improve protection. Instead, we trade a small amount of protection strength for lower LPIPS. This supports using these shallower prefix rather than using the full transformer.

\textbf{Objective direction.}
We finally isolate the direction of the attention objective. Both variants use the same double blocks, perturbation budget, and number of PGD steps; the only difference is whether target-to-reference attention is minimized or maximized. Minimizing reference attention barely changes edit success, from $81.6\%$ on the unprotected images to $80.7\%$. Maximizing the same attention instead reduces edit success to $5.1\%$, while also producing much larger drops in PickScore and CLIP$_{\mathrm{dir}}$. This matches the intervention result from Section~\ref{sec:intuition}. The important choice is not simply to manipulate reference attention, but to increase it enough to push the model toward copying.

\section{Conclusion}
Modern in-context image editors make heavy use of the reference image, so existing defenses have naturally tried to protect that image by making the model attend to it less. We find that this is often the wrong direction. Suppressing reference attention can leave the model free to follow the instruction, particularly when the protected subject occupies only a small fraction of the image. Controlled interventions reveal a different behavior when reference attention is increased. Once target tokens are made to rely strongly on the reference in the early transformer blocks, the model begins to reproduce the input instead of editing it. We turn this observation into \method, a protection objective that increases target-to-reference attention for each target token in a small set of early blocks. Across FLUX Kontext and FIBO Edit, \method provides stronger protection than existing baselines while maintaining comparable perturbation quality, and its early-layer objective makes optimization considerably cheaper.  Our results suggest that protecting images from modern editors need not mean hiding the reference from the model. Making the model rely on it more can itself be a defense.

\textbf{Limitations.} Our evaluation is limited to two DiT-based editors. Like other optimization-based proactive defenses, CaD also assumes white-box access to the protected editor during perturbation construction and requires per-image optimization, despite being substantially cheaper than deeper attention- or denoiser-based objectives. Finally, adversarial image protections can be weakened by common transformations such as compression, resizing, or noise, and by deliberate purification procedures; we do not evaluate robustness against such adaptive preprocessing or attacks here. 
\newpage

\subsection*{AI Use Statement}
In this work, we used generative AI tools to assist with implementation and debugging of methods and editing portions of the manuscript for clarity, structure, and presentation. In particular, AI tools were used to help implement models and attention intervention mechanisms and develop visualization code. We did not use generative AI to generate the benchmark datasets used in our experiments or to formulate or prove mathematical theorems; other required-disclosure tasks were not applicable to this work. All AI-assisted research code was reviewed by at least one author. Experimental results reported in the paper were obtained from our implemented evaluation pipeline rather than generated by an AI system, and at least one author visually inspected the resulting outputs and analyses.  We take responsibility for the final content of this work, including text, claims, code, and artifacts produced with the aid of generative AI.
\bibliography{iclr2027_conference}

\begin{thebibliography}{29}
\providecommand{\natexlab}[1]{#1}
\providecommand{\url}[1]{\texttt{#1}}
\expandafter\ifx\csname urlstyle\endcsname\relax
  \providecommand{\doi}[1]{doi: #1}\else
  \providecommand{\doi}{doi: \begingroup \urlstyle{rm}\Url}\fi

\bibitem[Bai et~al.(2025)Bai, Cai, Chen, Chen, et~al.]{bai2025qwen3vl}
Shuai Bai, Yuxuan Cai, Ruizhe Chen, Keqin Chen, et~al.
\newblock Qwen3-vl technical report.
\newblock \emph{arXiv preprint arXiv:2511.21631}, 2025.

\bibitem[{Black Forest Labs}(2025)]{labs2025kontext}
{Black Forest Labs}.
\newblock {FLUX.1 Kontext}: Flow matching for in-context image generation and editing in latent space.
\newblock \emph{arXiv preprint arXiv:2506.15742}, 2025.

\bibitem[Chen et~al.(2024)Chen, Yu, Ge, Yao, Xie, Wu, Wang, Kwok, Luo, Lu, and Li]{chen2024pixartalpha}
Junsong Chen, Jincheng Yu, Chongjian Ge, Lewei Yao, Enze Xie, Yue Wu, Zhongdao Wang, James Kwok, Ping Luo, Huchuan Lu, and Zhenguo Li.
\newblock {PixArt-$\alpha$}: Fast training of diffusion transformer for photorealistic text-to-image synthesis.
\newblock In \emph{International Conference on Learning Representations (ICLR)}, 2024.

\bibitem[Choi et~al.(2025)Choi, Lee, Jeong, Xie, Shin, and Lee]{choi2025diffusionguard}
June~Suk Choi, Kyungmin Lee, Jongheon Jeong, Saining Xie, Jinwoo Shin, and Kimin Lee.
\newblock {DiffusionGuard}: A robust defense against malicious diffusion-based image editing.
\newblock In \emph{International Conference on Learning Representations (ICLR)}, 2025.

\bibitem[Deng et~al.(2019)Deng, Guo, Xue, and Zafeiriou]{deng2019arcface}
Jiankang Deng, Jia Guo, Niannan Xue, and Stefanos Zafeiriou.
\newblock {ArcFace}: Additive angular margin loss for deep face recognition.
\newblock In \emph{IEEE/CVF Conference on Computer Vision and Pattern Recognition (CVPR)}, 2019.

\bibitem[Esser et~al.(2024)Esser, Kulal, Blattmann, Entezari, M{\"u}ller, Saini, Levi, Lorenz, Sauer, Boesel, Podell, Dockhorn, English, Lacey, Goodwin, Marek, and Rombach]{esser2024sd3}
Patrick Esser, Sumith Kulal, Andreas Blattmann, Rahim Entezari, Jonas M{\"u}ller, Harry Saini, Yam Levi, Dominik Lorenz, Axel Sauer, Frederic Boesel, Dustin Podell, Tim Dockhorn, Zion English, Kyle Lacey, Alex Goodwin, Yannik Marek, and Robin Rombach.
\newblock Scaling rectified flow transformers for high-resolution image synthesis.
\newblock In \emph{International Conference on Machine Learning (ICML)}, 2024.

\bibitem[Gal et~al.(2022)Gal, Patashnik, Maron, Bermano, Chechik, and Cohen-Or]{gal2022stylegannada}
Rinon Gal, Or~Patashnik, Haggai Maron, Amit~H. Bermano, Gal Chechik, and Daniel Cohen-Or.
\newblock {StyleGAN-NADA}: {CLIP}-guided domain adaptation of image generators.
\newblock \emph{ACM Transactions on Graphics (SIGGRAPH)}, 2022.

\bibitem[Ge et~al.(2026)Ge, Gandikota, Torralba, and Shaham]{ge2026vision}
Chris Ge, Rohit Gandikota, Antonio Torralba, and Tamar~Rott Shaham.
\newblock Vision-language binding in in-context image generation.
\newblock In \emph{Mechanistic Interpretability Workshop at ICML 2026}, 2026.

\bibitem[Gutflaish et~al.(2025)Gutflaish, Kachlon, Zisman, Hacham, Sarid, Visheratin, Huberman, Davidi, Bukchin, Goldberg, et~al.]{gutflaish2025generating}
Eyal Gutflaish, Eliran Kachlon, Hezi Zisman, Tal Hacham, Nimrod Sarid, Alexander Visheratin, Saar Huberman, Gal Davidi, Guy Bukchin, Kfir Goldberg, et~al.
\newblock Generating an image from 1,000 words: Enhancing text-to-image with structured captions.
\newblock \emph{arXiv preprint arXiv:2511.06876}, 2025.

\bibitem[H{\"o}nig et~al.(2025)H{\"o}nig, Rando, Carlini, and Tram{\`e}r]{honig2025adversarial}
Robert H{\"o}nig, Javier Rando, Nicholas Carlini, and Florian Tram{\`e}r.
\newblock Adversarial perturbations cannot reliably protect artists from generative ai.
\newblock In \emph{International Conference on Learning Representations}, volume 2025, pp.\  70223--70263, 2025.
\newblock URL \url{https://openreview.net/forum?id=yfW1x7uBS5}.

\bibitem[Jeon et~al.(2025)Jeon, Kim, Ha, Son, and Yoon]{jeon2025advpaint}
Joonsung Jeon, Woo~Jae Kim, Suhyeon Ha, Sooel Son, and Sung-eui Yoon.
\newblock {AdvPaint}: Protecting images from inpainting manipulation via adversarial attention disruption.
\newblock In \emph{International Conference on Learning Representations (ICLR)}, 2025.

\bibitem[Kirstain et~al.(2023)Kirstain, Polyak, Singer, Matiana, Penna, and Levy]{kirstain2023pickapic}
Yuval Kirstain, Adam Polyak, Uriel Singer, Shahbuland Matiana, Joe Penna, and Omer Levy.
\newblock Pick-a-pic: An open dataset of user preferences for text-to-image generation.
\newblock In \emph{Advances in Neural Information Processing Systems (NeurIPS)}, 2023.

\bibitem[Le et~al.(2023)Le, Phung, Nguyen, Dao, Tran, and Tran]{le2023antidreambooth}
Thanh~Van Le, Hao Phung, Thuan~Hoang Nguyen, Quan Dao, Ngoc~N. Tran, and Anh Tran.
\newblock Anti-{DreamBooth}: Protecting users from personalized text-to-image synthesis.
\newblock In \emph{IEEE/CVF International Conference on Computer Vision (ICCV)}, 2023.

\bibitem[Li et~al.(2026)Li, Abdel~Magid, Russakovsky, Gilbert, and Schoenebeck]{bodybench}
Qiwei Li, Salma Abdel~Magid, Olga Russakovsky, Eric Gilbert, and Sarita Schoenebeck.
\newblock Bodybench: Evaluating adversarial image defenses against ai nudification inpainting.
\newblock In \emph{Advances in Neural Information Processing Systems (NeurIPS)}, 2026.

\bibitem[Liang \& Wu(2023)Liang and Wu]{liang2023mist}
Chumeng Liang and Xiaoyu Wu.
\newblock Mist: Towards improved adversarial examples for diffusion models, 2023.

\bibitem[Liang et~al.(2023)Liang, Wu, Hua, Zhang, Xue, Song, Xue, Ma, and Guan]{liang2023advdm}
Chumeng Liang, Xiaoyu Wu, Yang Hua, Jiaru Zhang, Yiming Xue, Tao Song, Zhengui Xue, Ruhui Ma, and Haibing Guan.
\newblock Adversarial example does good: Preventing painting imitation from diffusion models via adversarial examples.
\newblock In \emph{International Conference on Machine Learning (ICML)}, 2023.

\bibitem[Liu et~al.(2025)Liu, Han, Xing, Yin, Wang, Cheng, Liao, Wang, Fu, Han, Li, Peng, Sun, Wu, Cai, Ge, Ming, Xia, Zeng, Zhu, Jiao, Zhang, Yu, and Jiang]{liu2025step1x}
Shiyu Liu, Yucheng Han, Peng Xing, Fukun Yin, Rui Wang, Wei Cheng, Jiaqi Liao, Yingming Wang, Honghao Fu, Chunrui Han, Guopeng Li, Yuang Peng, Quan Sun, Jingwei Wu, Yan Cai, Zheng Ge, Ranchen Ming, Lei Xia, Xianfang Zeng, Yibo Zhu, Binxing Jiao, Xiangyu Zhang, Gang Yu, and Daxin Jiang.
\newblock {Step1X-Edit}: A practical framework for general image editing.
\newblock \emph{arXiv preprint arXiv:2504.17761}, 2025.

\bibitem[Liu et~al.(2024)Liu, Fan, Dai, Chen, Zhou, and Sun]{liu2024metacloak}
Yixin Liu, Chenrui Fan, Yutong Dai, Xun Chen, Pan Zhou, and Lichao Sun.
\newblock {MetaCloak}: Preventing unauthorized subject-driven text-to-image diffusion-based synthesis via meta-learning.
\newblock In \emph{IEEE/CVF Conference on Computer Vision and Pattern Recognition (CVPR)}, pp.\  24219--24228, 2024.

\bibitem[Ma et~al.(2024)Ma, Goldstein, Albergo, Boffi, Vanden-Eijnden, and Xie]{ma2024sit}
Nanye Ma, Mark Goldstein, Michael~S. Albergo, Nicholas~M. Boffi, Eric Vanden-Eijnden, and Saining Xie.
\newblock {SiT}: Exploring flow and diffusion-based generative models with scalable interpolant transformers.
\newblock In \emph{European Conference on Computer Vision (ECCV)}, 2024.

\bibitem[Marchal et~al.(2024)Marchal, Xu, Elasmar, Gabriel, Goldberg, and Isaac]{marchal2024misuse}
Nahema Marchal, Rachel Xu, Rasmi Elasmar, Iason Gabriel, Beth Goldberg, and William Isaac.
\newblock Generative {AI} misuse: A taxonomy of tactics and insights from real-world data, 2024.
\newblock URL \url{https://arxiv.org/abs/2406.13843}.

\bibitem[{NBC News}(2026)]{nbc2026grok}
{NBC News}.
\newblock Musk's {AI} chatbot {Grok} is still making sexual deepfakes.
\newblock \url{https://www.nbcnews.com/tech/tech-news/musks-ai-chatbot-grok-xai-making-sexual-deepfakes-imagine-rcna265855}, April 2026.
\newblock Accessed: 2026-09-24.

\bibitem[Peebles \& Xie(2023)Peebles and Xie]{peebles2023dit}
William Peebles and Saining Xie.
\newblock Scalable diffusion models with transformers.
\newblock In \emph{IEEE/CVF International Conference on Computer Vision (ICCV)}, pp.\  4195--4205, 2023.

\bibitem[Radiya-Dixit et~al.(2021)Radiya-Dixit, Hong, Carlini, and Tramer]{radiya2021data}
Evani Radiya-Dixit, Sanghyun Hong, Nicholas Carlini, and Florian Tramer.
\newblock Data poisoning won't save you from facial recognition.
\newblock \emph{arXiv preprint arXiv:2106.14851}, 2021.

\bibitem[Ruiz et~al.(2020)Ruiz, Bargal, and Sclaroff]{ruiz2020disrupting}
Nataniel Ruiz, Sarah~Adel Bargal, and Stan Sclaroff.
\newblock Disrupting deepfakes: Adversarial attacks against conditional image translation networks and facial manipulation systems.
\newblock In \emph{European conference on computer vision}, pp.\  236--251. Springer, 2020.

\bibitem[Salman et~al.(2023)Salman, Khaddaj, Leclerc, Ilyas, and Madry]{salman2023photoguard}
Hadi Salman, Alaa Khaddaj, Guillaume Leclerc, Andrew Ilyas, and Aleksander Madry.
\newblock Raising the cost of malicious {AI}-powered image editing.
\newblock In \emph{Proceedings of the 40th International Conference on Machine Learning}, pp.\  29894--29918, 2023.

\bibitem[Shan et~al.(2023)Shan, Cryan, Wenger, Zheng, Hanocka, and Zhao]{shan2023glaze}
Shawn Shan, Jenna Cryan, Emily Wenger, Haitao Zheng, Rana Hanocka, and Ben~Y. Zhao.
\newblock Glaze: Protecting artists from style mimicry by text-to-image models.
\newblock In \emph{USENIX Security Symposium}, 2023.

\bibitem[Shen et~al.(2025)Shen, Cui, and Yang]{shen2025decontext}
Linghui Shen, Mingyue Cui, and Xingyi Yang.
\newblock {DeContext} as defense: Safe image editing in diffusion transformers.
\newblock \emph{arXiv preprint arXiv:2512.16625}, 2025.

\bibitem[Zhang et~al.(2018)Zhang, Isola, Efros, Shechtman, and Wang]{zhang2018lpips}
Richard Zhang, Phillip Isola, Alexei~A. Efros, Eli Shechtman, and Oliver Wang.
\newblock The unreasonable effectiveness of deep features as a perceptual metric.
\newblock In \emph{IEEE/CVF Conference on Computer Vision and Pattern Recognition (CVPR)}, 2018.

\bibitem[Zheng et~al.(2025)Zheng, Liang, and Wu]{zheng2025mistv2}
Boyang Zheng, Chumeng Liang, and Xiaoyu Wu.
\newblock Targeted attack improves protection against unauthorized diffusion customization.
\newblock In \emph{International Conference on Learning Representations (ICLR)}, 2025.

\end{thebibliography}
\bibliographystyle{iclr2027_conference}

\appendix

\section{Appendix}

\section{Protection Methods and Their Instantiation on Each Editing Model}
\label{sec:attack-methods}

We compare four protection methods (PhotoGuard, DeContext, Diff-PGD, and ours) on two image-editing models, FLUX.1~Kontext-dev and FIBO~Edit.
All methods perturb the \emph{context/reference} (to-be-protected) image $x$ with the same projected gradient descent (PGD) procedure and differ only in the loss they optimize.

\paragraph{Common PGD setup.}
Images are resized and center-cropped to $512\times512$ and normalized to $[-1,1]$.
The perturbed image $x'$ is kept in \texttt{float32} while the models run in \texttt{bfloat16}, so the perturbation is not lost to quantization.
Each step performs a signed-gradient update followed by projection onto the $\ell_\infty$ ball of radius $\varepsilon$ around $x$ and onto the valid image range:
\begin{equation}
  x' \leftarrow \operatorname{clip}_{[-1,1]}\!\Big(x + \operatorname{clip}_{[-\varepsilon,\varepsilon]}\big(x' + \alpha\,\operatorname{sign}(\nabla_{x'}\mathcal{L}) - x\big)\Big),
  \label{eq:pgd}
\end{equation}
with $\alpha=0.005$ in the $[-1,1]$ scale.
We use 800 steps for PhotoGuard, DeContext and Diff-PGD, and 1000 for our method. The sign in \eqref{eq:pgd} is $+$ when we ascend a loss and $-$ when we descend one; each method below states which.

\paragraph{Notation.}
$E$ is the model's VAE encoder (posterior mode/mean), $F_\theta$ its diffusion transformer, $p$ an edit instruction from a pool $\mathcal{P}$ of 60 instructions directly from \cite{shen2025decontext} (expressions, accessories, poses, style and scene changes, and combinations), and $t$ a timestep drawn uniformly from $\{980,\dots,999\}$, \textit{i.e.}\ nearly pure noise.
Both models are flow-matching models with noisy latent $z_t=(1-\sigma_t)z+\sigma_t n$, $n\sim\mathcal{N}(0,I)$.
In both models the transformer processes a joint sequence of tokens
$[\,T \mid Z \mid C\,]$: text tokens $T$, target (to-be-generated) image tokens $Z$, and context image tokens $C=\operatorname{tokens}(E(x'))$.
At $512\times512$ each image stream has $N=1024$ tokens.

\subsection{Model-specific components}

\paragraph{FLUX.1~Kontext-dev.}
The VAE is a 16-channel KL autoencoder with $8\times$ spatial downsampling; latents are shifted and scaled by scalar factors and packed into $2\times2$ patches.
Text conditioning is a pooled CLIP embedding plus 512 T5 tokens.

\paragraph{FIBO~Edit.}
The VAE is the Wan~2.2 video VAE (48 latent channels, $16\times$ downsampling, no patch packing), applied to a single frame; latents are normalized per channel, $z=(z_{\text{raw}}-\mu)/\sigma$.
Text conditioning uses a SmolLM3 encoder, whose hidden states and a per-block text layer are injected into the transformer, together with a joint additive attention mask over $[T\mid Z\mid C]$; the text length varies with the prompt and there is no guidance embedding.
FIBO~Edit expects a structured JSON prompt, so for each (image, instruction) pair we obtain a JSON description from the FIBO prompt-to-JSON vision-language model (VLM) applied to the \emph{clean} image, and cache it.

\subsection{PhotoGuard (encoder attack)}
PhotoGuard attacks only the VAE encoder and does not involve the transformer or any prompt.
It starts from a random point $x'_0=\operatorname{clip}(x+u)$, $u\sim\mathcal{U}(-\varepsilon,\varepsilon)$, and \emph{descends}
\begin{equation}
  \mathcal{L}_{\text{PG}}(x') = \big\lVert E(x') - E(x_{\text{gray}})\big\rVert_2,
\end{equation}
which pushes the encoding of $x'$ towards that of a mid-gray image $x_{\text{gray}}=0$.
The step size decays linearly from $\alpha$ to $\alpha/100$ over the run.
\emph{FLUX:} $E$ is the FLUX KL-VAE. \emph{FIBO:} $E$ is the Wan VAE.

\subsection{DeContext (context-attention suppression)}
DeContext ~\citep{shen2025decontext} lowers the attention that generated tokens pay to the context image.
At each step we sample $p\in\mathcal{P}$ and $t$, feed the transformer the packed sequence of pure noise for the target stream and the tokens of $E(x')$ for the context stream, and hook the query/key tensors after the QK-normalization of the first $26$ single-stream blocks $\mathcal{S}$.
After applying rotary embeddings, the attention of target queries over all keys is
\begin{equation}
  A^{(l,h)} = \operatorname{softmax}\!\Big(\tfrac{Q_Z^{(l,h)}\,K^{(l,h)\top}}{\sqrt{d}} \,[+\,M]\Big),
\end{equation}
and the context proportion of layer $l$ is $m_C^{(l)}=\operatorname{mean}_{h,q}\sum_{j\in C}A^{(l,h)}_{q,j}$.
We \emph{ascend}
\begin{equation}
  \mathcal{L}_{\text{DC}}(x') = 1-\frac{1}{|\mathcal{S}|}\sum_{l\in\mathcal{S}} m_C^{(l)},
\end{equation}
\textit{i.e.}\ we minimize the mean context attention proportion.
\emph{FLUX:} the sequence is $512+1024+1024=2560$ tokens (T5 length fixed at 512).
\emph{FIBO:} the same objective is used, but the text length is read from the current prompt instead of being fixed, and the additive attention mask $M$ (masking padded text tokens) is added to the scores before the softmax so that the softmax matches the model's own.

\subsection{Diff-PGD (edit-reconstruction loss)}
Diff-PGD, proposed by ~\citep{shen2025decontext}, as a sensible baseline maximizes the flow-matching error of the model when it tries to produce a clean edit from the protected context.
For each $(x,p)$ we first generate a clean edited target $y_p=G(x,p)$ with the editing model itself, always from the original $x$, and encode it, $z^\star_p=E(y_p)$.
Targets are generated lazily the first time a prompt is sampled and cached on disk and in memory, so each prompt costs one generation per image.
At each step we sample $p$, $t$ and noise $n$, form $z_t=(1-\sigma_t)z^\star_p+\sigma_t n$, and \emph{ascend}
\begin{equation}
  \mathcal{L}_{\text{DP}}(x') = \Big\lVert \big[F_\theta\big([\,z_t \mid \operatorname{tokens}(E(x'))\,],\,p,\,t\big)\big]_{Z} - (n - z^\star_p)\Big\rVert_2^2 ,
\end{equation}
where $[\cdot]_Z$ keeps the outputs at the target-stream tokens and $n-z^\star_p$ is the flow-matching velocity target.
\emph{FLUX:} $G$ is FLUX Kontext run for 28 steps with guidance scale $3.5$, using the precomputed prompt embeddings and a separate scheduler instance so the attack's scheduler is unaffected.
\emph{FIBO:} $G$ is the FIBO~Edit pipeline run for 30 steps with classifier-free guidance scale $5.0$
The JSON prompt (VLM-generated by default) is used both to generate the target and in the loss, and VLM-JSON targets are cached separately from those made with the minimal prompt.

\subsection{Ours (context-attention maximization)}
Our method also uses the attention of the target stream but acts on the double-stream (joint) blocks and drives the target queries' attention mass \emph{onto} the reference (context) tokens.
No prompt pool is needed: the pool contains a single empty prompt, so the attack is unconditional.
For each selected double block $l\in\mathcal{D}$ we hook the normalized queries and keys of both streams (image stream $[Z\mid C]$ and text stream $T$), rebuild the joint sequence $[T\mid Z\mid C]$, apply the joint rotary embedding, and compute the ordinary softmax attention of the target queries.
With $m_C^{(l)}$ as above (here called the reference mass, computed \emph{per query} before averaging), we \emph{ascend}
\begin{equation}
  \mathcal{J}(x') = -\frac{1}{|\mathcal{D}|}\sum_{l\in\mathcal{D}}\ \operatorname{mean}_{h,q}\Big(1-\textstyle\sum_{j\in C}A^{(l,h)}_{q,j}\Big)^{2},
\end{equation}
which pushes the reference mass of every head and target query towards $1$.
\emph{FLUX:} $\mathcal{D}$ is double blocks $0$--$9$, and the empty prompt is encoded by CLIP and T5 (512 tokens) with guidance $3.5$.
\emph{FIBO:} $\mathcal{D}$ is double blocks $0$--$3$ (half of the eight FIBO~Edit's double blocks); the additive attention mask is added to the scores as in the model's own attention; and the empty instruction is turned into a single VLM-generated JSON per image, so the VLM is called once per image.

\end{document}